\documentclass[final]{cvpr}

\usepackage{times}
\usepackage{epsfig}
\usepackage{graphicx}
\usepackage{amsmath}
\usepackage{amssymb}
\usepackage{color}

\usepackage{comment}
\usepackage{amsmath,amssymb} 
\usepackage{multirow, array, booktabs,xspace,bm}
\usepackage{pifont}
\usepackage{amsfonts}
\usepackage{arydshln}
\usepackage{cite}
\usepackage{makecell}
\usepackage{dcolumn}
\usepackage{eqparbox}
\usepackage{subcaption}

\usepackage{algorithm,algpseudocode}

\newcolumntype{P}[1]{>{\centering\arraybackslash}p{#1}}
\newcolumntype{M}[1]{>{\centering\arraybackslash}m{#1}}

\usepackage[pagebackref=true,breaklinks=true,colorlinks,bookmarks=false]{hyperref}

\def\cvprPaperID{7876} 
\def\confYear{CVPR 2021}

\definecolor{red2}{rgb}{1.0,0.20,0.20}
\definecolor{blue2}{rgb}{0.220,0.407,0.82}
\begin{document}

\title{Video Prediction Recalling Long-term Motion Context \\ via Memory Alignment Learning}

\author{Sangmin Lee$^{1}$ \;\; Hak Gu Kim$^{2}$ \;\; Dae Hwi Choi$^{1}$ \;\; Hyung-Il Kim$^{1, 3}$ \;\; Yong Man Ro$^{1}$\thanks{Corresponding author}\\ 
$^{1}$ Image and Video Systems Lab, KAIST, South Korea\\
$^{2}$ EPFL, Switzerland  \;\;\;  $^{3}$ ETRI, South Korea\\
{\tt\small\{sangmin.lee,sinjh1796,ymro\}@kaist.ac.kr \ hakgu.kim@epfl.ch \ hikim@etri.re.kr}
}

\maketitle

\begin{abstract}
   Our work addresses long-term motion context issues for predicting future frames. To predict the future precisely, it is required to capture which long-term motion context (\textit{e.g.}, walking or running) the input motion (\textit{e.g.}, leg movement) belongs to. The bottlenecks arising when dealing with the long-term motion context are: (i) how to predict the long-term motion context naturally matching input sequences with limited dynamics, (ii) how to predict the long-term motion context with high-dimensionality (\textit{e.g.}, complex motion). To address the issues, we propose novel motion context-aware video prediction. To solve the bottleneck (i), we introduce a long-term motion context memory (LMC-Memory) with memory alignment learning. The proposed memory alignment learning enables to store long-term motion contexts into the memory and to match them with sequences including limited dynamics. As a result, the long-term context can be recalled from the limited input sequence. In addition, to resolve the bottleneck (ii), we propose memory query decomposition to store local motion context (\textit{i.e.}, low-dimensional dynamics) and recall the suitable local context for each local part of the input individually. It enables to boost the alignment effects of the memory. Experimental results show that the proposed method outperforms other sophisticated RNN-based methods, especially in long-term condition. Further, we validate the effectiveness of the proposed network designs by conducting ablation studies and memory feature analysis. The source code of this work is available\footnote[2]{\url{https://github.com/sangmin-git/LMC-Memory}}. 
\end{abstract}

\section{Introduction}
Video prediction in computer vision is to estimate upcoming future frames at pixel-level from given previous frames. Since predicting the future is an important basement for intelligent decision-making systems, the video prediction has attracted increasing attention in industry and research fields. It has the potential to be applied to various tasks such as weather forecasting \cite{xingjian2015convolutional}, traffic situation prediction \cite{chandra2019traphic}, and autonomous driving \cite{castrejon2019improved}. However, the pixel-level video prediction is still challenging mainly due to the difficulties of capturing high-dimensionality and long-term motion dynamics \cite{villegas2017learning, villegas2017decomposing, hsieh2018learning, wang2018eidetic}.

Recently, several studies with deep neural networks (DNNs) have been proposed to capture the high-dimensionality and the long-term dynamics of video data in the video prediction field \cite{finn2016unsupervised, villegas2017decomposing, hsieh2018learning, villegas2017learning, wang2018predrnn++, wang2018eidetic, su2020convolutional}. The models considering the high-dimensionality of videos tried to simplify the problem by constraining motion and disentangling components \cite{finn2016unsupervised, villegas2017decomposing,hsieh2018learning}. However, these methods did not consider the long-term frame dynamics, which leads to predicting blurry frames or wrong motion trajectories. Recurrent neural networks (RNNs) have been developed to capture the long-term dynamics with consideration for long-term dependencies in the video prediction \cite{villegas2017learning, wang2018predrnn++, wang2018eidetic}. The long-term dependencies in the RNNs is about remembering past step inputs. The RNN-based methods exploited the memory cell states in the RNN unit. The cell states are recurrently changed according to the current input sequence to remember the previous steps of the sequence. However, it is difficult to capture the long-term motion dynamics for the input sequence with limited dynamics (\textit{i.e.}, short-term motion) because such cell states mainly depend on revealing relations within the current input sequence. For example, given short-length input frames for a walking motion, the leg movement from the input is limited itself. Therefore, it is difficult to grasp what will happen to the leg in the future through the cell states of the RNNs. In this case, the long-term motion context of the partial action may not be properly captured by the RNN-based methods.

Our work addresses \textit{long-term motion context issues} for predicting future frames, which have not been properly dealt with in previous video prediction works. To predict the future precisely, it is required to capture which long-term motion context the input motion belongs to. For example, in order to predict the future of leg movement, we need to know such partial leg movement belongs to either walking or running (\textit{i.e.}, long-term motion context). The bottlenecks arising when dealing with long-term motion context are as follows: \textit{(i) how to predict the long-term motion context naturally matching input sequences with limited dynamics, (ii) how to predict the long-term motion context with high-dimensionality}.

In this paper, we propose novel motion context-aware video prediction to address the aforementioned issues. To solve the bottleneck \textit{(i)}, we introduce a long-term motion context memory (LMC-Memory) with memory alignment learning. Contrary to the internal memory cells of the RNNs, the LMC-Memory externally exists with its own parameters to preserve various long-term motion contexts of training data, which are not limited to the current input. Memory alignment learning is proposed to effectively store long-term motion contexts into the LMC-Memory and recall them even with inputs having limited dynamics. Memory alignment learning contains two training phases to align long-term and short-term motions: $\left<\text{\textit{Phase 1}}\right>$ storing long-term motion context from long-term sequences into the memory, $\left<\text{\textit{Phase 2}}\right>$ matching input short-term sequences with the stored long-term motion contexts in the memory. As a result, the long-term motion context (\textit{e.g.}, long-term walking dynamics) can be recalled from the input short-term sequence alone (\textit{e.g.}, short-term walking clip).

Furthermore, to resolve the bottleneck \textit{(ii)}, we propose decomposition of a memory query that is used to store and recall the motion context. Even if various motion contexts of training data are stored in the LMC-Memory, it is difficult to capture the motion context that is exactly matched with the input. This is because motions of video sequences have high-dimensionality (\textit{e.g.}, complex motion with local motion components). The dimensionality indicates the number of pixels in a video sequence. Since each motion is slightly different from one another in a global manner even for the same category, the proposed memory query decomposition is useful in that it enables to store local context (\textit{i.e.}, low-dimensional dynamics) and recall the suitable local context for each local part of the input individually. It can boost the alignment effects between the input and the stored long-term motion context in the LMC-Memory.

The major contributions of the paper are as follows.
\vspace{-0.3cm}
\begin{itemize}
	\setlength\itemsep{-0.3em}
	\item We introduce novel motion context-aware video prediction to solve the inherent problem of the RNN-based methods in capturing long-term motion context. We address the arising long-term motion context issues in the video prediction.
	\item	We propose the LMC-Memory with memory alignment learning to address storing and recalling long-term motion contexts. Through the learning, it is possible to recall long-term motion context corresponding to an input sequence even with limited dynamics.
	\item To address the high-dimensionality of motions, we decompose memory query to separate an overall motion into local motions with low-dimensional dynamics. It makes it possible to recall suitable local motion context for each local part of the input individually.
\end{itemize}

\section{Related Work}
\subsection{Video Prediction}
In video prediction, errors for predicting future frames can be divided into two factors \cite{wang2018eidetic}. The first one is about systematic errors due to the lack of modeling capacity for deterministic changes. The second one is related to modeling the intrinsic uncertainty of the future. There have been several works to address the second factor \cite{babaeizadeh2017stochastic, lee2018stochastic, denton2018stochastic, xu2020stochastic}. These methods utilized stochastic modeling to generate plausible multiple futures. Contrary to these, our paper addresses the video prediction focusing on the first factor. 

Recently, deep learning-based video prediction methods have been proposed to deal with the first factor. They considered the problems leading to prediction difficulty such as capturing high-dimensionality and long-term dynamics in video data \cite{patraucean2015spatio,finn2016unsupervised,villegas2017decomposing, villegas2017learning,kalchbrenner2017video,wang2017predrnn, villegas2018hierarchical,gao2019disentangling,ye2019compositional,xu2018video,byeon2018contextvp,wang2018predrnn++,wang2018eidetic,minderer2019unsupervised,villegas2019high,kim2019unsupervised,yu2020efficient, wu2020future, jin2020exploring, su2020convolutional}. Finn \textit{et al.} \cite{finn2016unsupervised} incorporated appearance information in the previous frames with the predicted pixel motion information for long-range video prediction. Villegas \textit{et al.} \cite{villegas2017learning} introduced a hierarchical prediction model that generates the future image from the predicted high-level structure. A predictive recurrent neural network (PredRNN) model was presented by Wang \textit{et al.} \cite{wang2017predrnn} to model and memorize both spatial and temporal representations simultaneously. Wang \textit{et al.}  \cite{wang2018predrnn++} further extended this model, named PredRNN++ to solve the vanishing gradient problem in deep-in-time prediction by building adaptive learning between long-term and short-term frame relation. Recently, eidetic 3D LSTM (E3D-LSTM) \cite{wang2018eidetic} was proposed  to integrate 3D convolutions into the RNNs for effectively addressing memories across long-term periods. Jin \textit{et al.} \cite{jin2020exploring} introduced spatial-temporal multi-frequency analysis for high-fidelity video prediction with temporal-consistency . Su \textit{et al.} \cite{su2020convolutional} proposed convolutional tensor-train decomposition to learn long-term spatio-temporal correlations.  However, these works still have a limitation in encoding long-term dynamics in that they mainly rely on the input sequence to find frame relations. Therefore, it is difficult to capture the long-term motion context for predicting the future from the input sequence with limited dynamics.

\begin{figure*}[t]
	\begin{minipage}[b]{1.0\linewidth}
		\centering
		\centerline{\includegraphics[width=14.5cm]{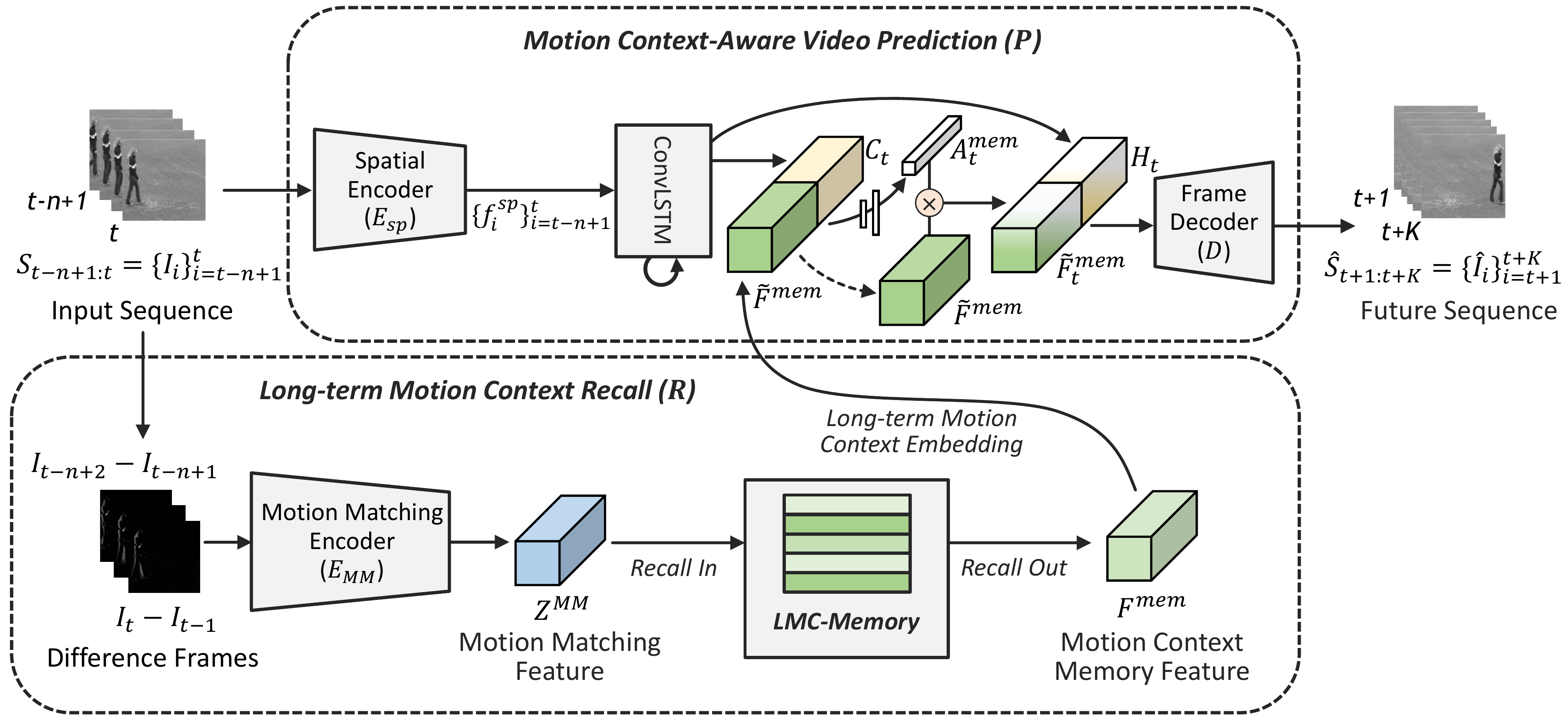}}
	\end{minipage}
		\vspace{-0.6cm}
	\caption{Overall framework with the proposed LMC-memory for video prediction at testing phase. The lower path is for recalling long-term motion context from the external memory, named LMC-Memory. The upper path is for predicting future frames with recurrent manner considering  the recalled long-term motion context.}
	\label{figure1}
		\vspace{-0.2cm}
\end{figure*}

\subsection{Memory Network}
Memory augmented networks have recently been introduced for solving various problems in computer vision tasks \cite{gong2019memorizing, kaiser2017learning,cai2018memory,zhu2019dm,lee2018memory,pei2019memory, zhu2020inflated,han2020memory,lai2020mast,park2020learning}. Such computer vision tasks include anomaly detection \cite{gong2019memorizing,park2020learning}, few-shot learning \cite{kaiser2017learning, cai2018memory, zhu2020inflated}, image generation \cite{zhu2019dm}, and video summarization \cite{lee2018memory}. Kaiser \textit{et al.} \cite{kaiser2017learning} presented a large scale long-term memory module for life-long learning. Memory-attended recurrent network was proposed by Pei \textit{et al.} \cite{pei2019memory} to capture the full-spectrum correspondence between the word and its visual contexts across video sequences in training data. To utilize the external memory network for our purposes, we introduce novel memory alignment learning that enables to store the long-term motion contexts into the memory and to recall them with limited input sequences. In addition, we separate overall motion into low-dimensional dynamics and utilize them as an individual memory query to recall proper long-term motion context for each local part of inputs.

\section{Proposed Method}
\subsection{Motion Context-Aware Video Prediction}
Video prediction task can be formulated as follows. Let ${I}_t$ $\in$ $\mathbb{R}^{W\times H\times C}$ denote the $(t)$-th frame in the video, and ${S}_{t-n+1:t}$ $=$ $\left\{{I}_{i}\right\}_{i=t-n+1}^{t}$ denote the video sequence containing $(t\text{-}n\text{+}1)$-th to $(t)$-th frames. The goal is to optimize the predictive function $\mathcal{F}$ for making generated next sequence $\hat{S}_{t+1:t+K}$$=$$\mathcal{F}({S}_{t-n+1:t})$ be similar with actual next sequence ${S}_{t+1:t+K}$ for given previous sequence ${S}_{t-n+1:t}$. Figure \ref{figure1} shows the overall framework of the proposed video frame prediction at inference phase. The input sequence goes through two paths to predict the future frames. One (lower path of Figure \ref{figure1}) is for recalling long-term motion context from the memory. The other (upper path of Figure \ref{figure1}) is for predicting frames recurrently with the recalled long-term motion context.  

First, in the lower path of Figure \ref{figure1}, the differences between the consecutive frames (\textit{i.e.}, difference frames) are used as inputs of motion matching encoder $E_{MM}$. Then, a motion matching feature $Z^{MM}$ is extracted to recall the motion context memory feature ${F}^{mem}$ from the external memory, named LMC-Memory. This LMC-Memory contains various long-term motion contexts of training data. Thus, ${F}^{mem}$ from the memory can be considered as long-term information corresponding to the input sequence ${S}_{t-n+1:t}$ (described in Section 3.2 in detail). It is then embedded in the upper part $P$. This long-term motion context embedding contains 2D-DeConvs to match the spatial size with the upper part, which results in $\widetilde{F}^{mem}$.

\begin{figure*}[t]
	\begin{minipage}[b]{1.0\linewidth}
		\centering
		\centerline{\includegraphics[width=17.5cm]{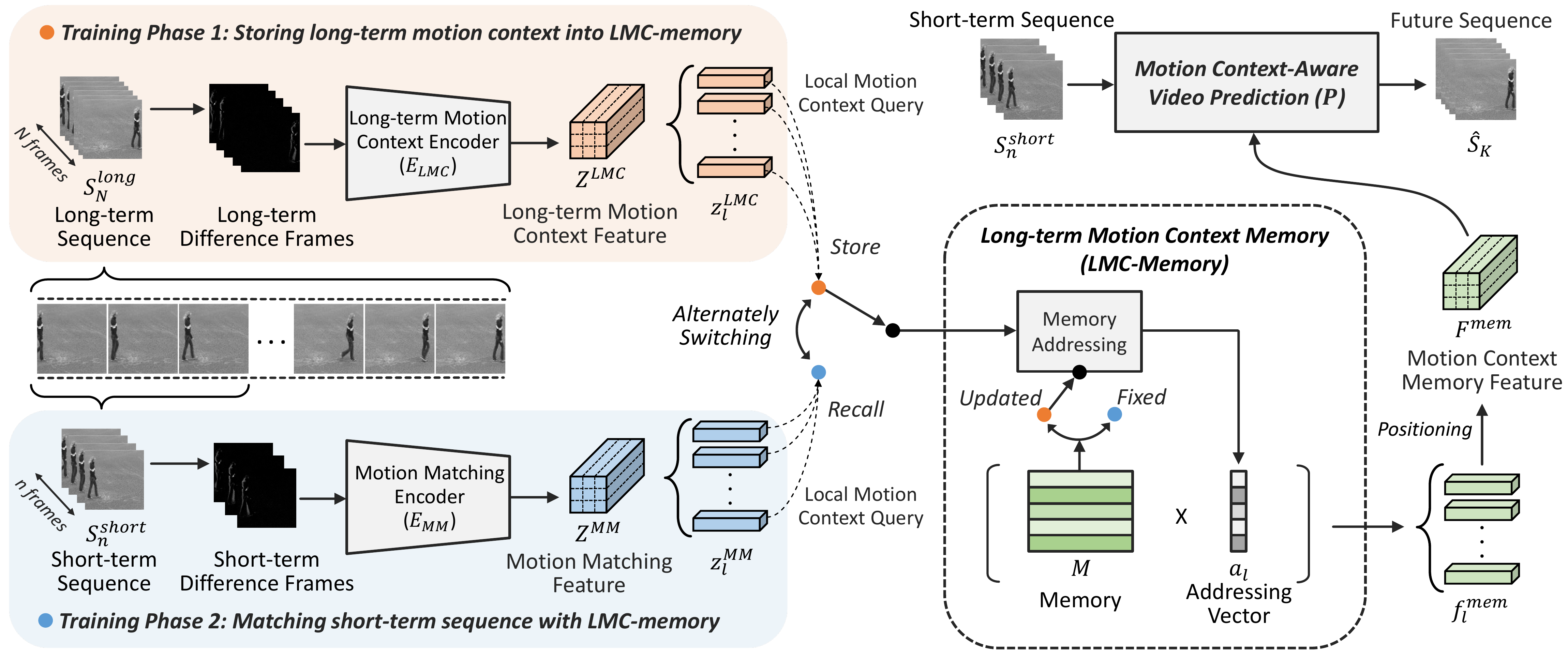}}
	\end{minipage}
	\vspace{-0.6cm}
	\caption{Training scheme of LMC-memory. To align the long-term and short-term in the memory, the networks are trained with two phases: (1) storing long-term motion context, (2) matching limited short-term sequence with the long-term context.}
	\label{figure2}
	\vspace{-0.2cm}
\end{figure*}

The upper part of Figure \ref{figure1}  demonstrates long-term motion context-aware video prediction $P$ scheme. In this path, the required motion context is refined through attention-based encoding to effectively embed it in predicting future frames. Each frame of the input sequence is independently fed to spatial encoder $E_{sp}$ with 2D-Convs to extract appearance characteristics. The ConvLSTM \cite{xingjian2015convolutional} receives each extracted spatial feature $f_{t}^{sp}=E_{sp}(I_{t})$ as inputs in time step order. A cell state $C_t$ and an output state $H_t$ are obtained from recurrent processing of the ConvLSTM. Since $C_t$ contains the information from the past to the present of the input sequence, we use $C_t$ to refine $\widetilde{F}^{mem}$ for embedding the required motion context at the current step. $C_t$ and $\widetilde{F}^{mem}$ are concatenated and pass through fully connected layers to make channel-wise attention $A^{mem}_{t}$ for $\widetilde{F}^{mem}$. The channel-wise refined feature ${{\widetilde{F}}}^{mem}_{t}$$=$$A^{mem}_{t}\otimes\widetilde{F}^{mem}$ and output state $H_t$ from the ConvLSTM are concatenated to embed long-term context to the ConvLSTM output (\textit{i.e.}, spatio-temporal information of the input). The concatenated feature is fed to a frame decoder $D$ with 2D-DeConvs to generate corresponding next frame $\hat{I}_{t+1}=D([H_t; {{\widetilde{F}}}^{mem}_{t}])$. The embedded motion context memory feature can provide the prior of long-term motion context for the current input sequence. Note that the generated next frame $\hat{I}_{t+1}$ enters $P$ as a new input to create the further future frame.

\subsection{LMC-Memory with Alignment Learning}
The long-term motion context memory, named LMC-Memory is to provide the long-term motion context for current input sequences to predict future frames. To effectively recall the long-term motion context even for the input sequence with limited dynamics, we propose novel memory alignment learning. Figure \ref{figure2} shows the training scheme of the LMC-Memory. The memory is trained alternately with two phases: $\left<\text{\textit{Phase 1}}\right>$ storing long-term motion context into the memory and $\left<\text{\textit{Phase 2}}\right>$ matching a limited sequence with the corresponding long-term context in the memory.

During the storing phase $\left<\text{\textit{Phase 1}}\right>$, we take a long-term sequence $S^{long}_{N}$ with length $N$ from the training data. After obtaining the difference frames, the motion context of the long-term sequence is extracted by a long-term motion context encoder $E_{LMC}$. We adopt typical motion extractor, C3D \cite{tran2015learning} with 3D-Convs for $E_{LMC}$. The resulting long-term motion context feature $Z^{LMC}=\left\{z^{LMC}_{l}\right\}_{l=1}^{w \times h}$$\in$$\mathbb{R}^{w\times h\times c}$ is divided into local parts to exploit decomposed dynamics. The local feature $z^{LMC}_{l}$$\in$$\mathbb{R}^{c}$ is used as a memory query individually.

The parameters of the LMC-Memory have a matrix form, $M=\left\{m_{i}\right\}_{i=1}^{s}$$\in$$\mathbb{R}^{s\times c}$ with $s$ slot size and $c$ channels. A row vector $m_{i}$$\in$$\mathbb{R}^{c}$ denotes a memory item of $M$. An addressing vector $a_l=\left\{a_{l\_i}\right\}_{i=1}^{s}$$\in$$\mathbb{R}^{s}$ for query $z^{LMC}_{l}$ is used to address the location of the memory $M$. Each scalar value $a_{l\_i}$ of $a_l$ can be considered as an attention weight for the corresponding memory slot $m_{i}$. Memory addressing procedure can be formulated as
\begin{equation}
a_{l\_i} = \frac{\exp(d(z^{LMC}_{l}, m_{i}))}{\sum_{j=1}^{s} \exp(d(z^{LMC}_{l}, m_j))},
\label{eq:1}
\end{equation}
where $d(\cdot,\cdot)$ indicates cosine similarity function and $\exp(\cdot)/\sum \exp(\cdot)$ denotes softmax function. With $M$ and $a_l=\left\{a_{l\_i}\right\}_{i=1}^{s}$, the memory outputs a local motion context memory feature ${f}^{mem}_{l}$$\in$$\mathbb{R}^{c}$ $(l=1, 2, ..., w \times h)$ for each location $l$ as follows 
\begin{equation}
{f}^{mem}_{l} = \sum\nolimits_{i=1}^s a_{l\_i} m_{i}.
\label{eq:2}
\end{equation}
Finally, a motion context memory feature ${F}^{mem}=\left\{{f}^{mem}_{l}\right\}_{l=1}^{w \times h}$$\in$$\mathbb{R}^{w\times h\times c}$ is obtained by positioning each local feature ${f}^{mem}_{l}$ as shown in Figure \ref{figure2}. As addressed in Section 3.1, ${F}^{mem}$ is embedded to motion context-aware video prediction $P$. During the training phase 1, the weights of the memory $M$ are updated through backpropagation as \cite{gong2019memorizing}. We train the networks to generate long-term future from long-term input so that long-term motion context can be stored in the memory at this phase.

At the matching phase $\left<\text{\textit{Phase 2}}\right>$, the model receives a short-term sequence $S^{short}_{n}$ with length $n$ (long-term length $N$ $>$ short-term length $n$). The matching phase allows long-term information in the memory to be recalled by a limited short-term sequence. Similar to the long-term encoding process, the difference frames are utilized for motion encoding. Then, motion matching feature $Z^{MM}$ is extracted by a motion matching encoder $E_{MM}$. Same as $E_{LMC}$, $E_{MM}$ has the C3D \cite{tran2015learning} structure, but does not share parameters with $E_{LMC}$. The local feature $z^{MM}_{l}$ of $Z^{MM}=\left\{z^{MM}_{l}\right\}_{l=1}^{w \times h}$$\in$$\mathbb{R}^{w\times h\times c}$ is  used to recall the corresponding long-term motion context from the memory. The memory addressing procedures are the same as the first storing phase (Eq.\ \ref{eq:1} and \ref{eq:2}). However, unlike the phase 1, the weights of the memory $M$ are not optimized and only used to recall the motion context during this matching phase. This is to preserve the stored long-term motion context in $M$. Except for the memory $M$, overall network weights are trained to predict the long-term future frames with the memory feature. Thus, it enables $E_{MM}$ to extract $Z^{MM}$ that properly recalls the corresponding long-term motion context in the given LMC-memory. 

\begin{algorithm}[t!]
	\caption{Memory Alignment Learning} \label{alg1}
	\begin{algorithmic}[1]
		\State {\bf Inputs}: short-term sequence $S^{short}_{n}=S_{t-n+1:t}$, long-term sequence $S^{long}_{N}=S_{t-n+1+r:t-n+r+N}$ (random integer $r \sim \mathcal{U}\{0, n\}$), and learning rate $\alpha$.
		
		\State Initialize parameters of \textsc{Phase 1} networks ($\theta$), \textsc{Phase 2} networks ($\phi$), and LMC-memory ($M$). 
		\vspace{0.05in}
		\hrule
		\vspace{0.05in}
		\For{each iteration}
		\State {$\left\langle {\textsc{\textbf{Phase 1}: Storing Phase}}\right\rangle$}
		\State Get  $Z^{LMC}=E_{LMC}(S^{long}_{N})$
		\State Get  $F^{mem}=LMC\text{-}Memory(Z^{LMC})$
		\For{$i=0,1,..., K\text{-}1$}
		\State Get $\hat{I}_{t+i+1}=P(S_{t-n+1:t}, \hat{S}_{t+1:t+i}, {F}^{mem})$
		\EndFor
		\State $\mathcal{L} \leftarrow \mathcal{L}^{\tt pred}(\hat{S}_{t+1:t+K}, {S}_{t+1:t+K})$
		\State Update
		$\theta \leftarrow \theta - \alpha \nabla_{\theta} \mathcal{L}$
		\State {$\left\langle {\textsc{\textbf{Phase 2}: Matching Phase}}\right\rangle$}
		\State Get  $Z^{MM}=E_{MM}(S^{short}_{n})$
		\State Get  $F^{mem}=LMC\text{-}Memory(Z^{MM})$
		\For{$i=0,1,..., K\text{-}1$}
		\State Get $\hat{I}_{t+i+1}=P(S_{t-n+1:t}, \hat{S}_{t+1:t+i}, {F}^{mem})$
		\EndFor
		\State $\mathcal{L} \leftarrow \mathcal{L}^{\tt pred}(\hat{S}_{t+1:t+K}, {S}_{t+1:t+K})$
		\State Update $\phi \ (except \ M) \leftarrow \phi - \alpha \nabla_{\phi} \mathcal{L}$
		
		\EndFor
	\end{algorithmic}
\end{algorithm}

Optimization is performed with the prediction framework $P$ (see Figure \ref{figure2}). Only the short-term sequence $S^{short}_{n}={S}_{t-n+1:t}$ is fed as a main input of $P$. The memory path receives the long-term $S^{long}_{N}$ and short-term $S^{short}_{n}$ alternately. Two training phases are alternately performed in each iteration. In both phases, according to \cite{wang2017predrnn, wang2018predrnn++, wang2018eidetic, su2020convolutional}, we exploit a prediction loss function $\mathcal{L}^{\tt pred}$  as follows
\begin{equation}
\begin{split}
\mathcal{L}^{\tt pred}=&\|\hat{S}_{t+1:t+K}-S_{t+1:t+K}\|^2_2 \\
&+\|\hat{S}_{t+1:t+K}-S_{t+1:t+K}\|_1,
\label{eq:3}
\end{split}
\end{equation}
where $\hat{S}_{t+1:t+K}$ denotes $K$ predicted future frames while $S_{t+1:t+K}$ denotes $K$ ground truth future frames. Note that the proposed method only takes short-term sequences at inference time as shown in Figure \ref{figure1}. Training procedure is further described in Algorithm \ref{alg1}. 


\begin{table*}
	\renewcommand{\arraystretch}{1.0}
	\renewcommand{\tabcolsep}{4.1mm}
	
	\centering
	
	\resizebox{0.98\linewidth}{!}{
		\begin{tabular}{l  c c c  c c c c }
			\toprule
			\multirow{2}{*}{\bf Prediction Method} & \multicolumn{3}{c}{\bf \makecell{Performance\\(10 $\to$ 10)}} & \multicolumn{3}{c}{\bf \makecell{Performance\\(10 $\to$ 30)}} & \bf \makecell{Computational Cost\\(10 $\to$ 30)} \\
			\cmidrule(lr){2-4}
			\cmidrule(lr){5-7}
			
			& \bf MSE & \bf SSIM & \bf LPIPS &\bf  MSE & \bf SSIM & \bf LPIPS & \bf Inference Time (s) \\
			\bottomrule
			\rule{0pt}{10pt}TRAJGRU~\cite{shi2017deep}
			& 106.9 & 0.713 & - & 163.0& 0.588 & - &-  \\
			CDNA~\cite{finn2016unsupervised}
			& 97.4 & 0.721 & - & 142.3 & 0.609 & - &-  \\
			VPN~\cite{kalchbrenner2017video}
			& 64.1 & 0.870 & - & 129.6 & 0.620 & -  &- \\
			PredRNN~\cite{wang2017predrnn}
			& 56.8 & 0.867 & - & 112.2 & 0.645 & - &- \\
			PredRNN++~\cite{wang2018predrnn++} & \bf\textcolor{blue2}{42.1} & 0.913 &  59.5 & \bf\textcolor{blue2}{84.0} & 0.834 & 139.9  &0.308
			\\
			E3D-LSTM~\cite{wang2018eidetic} & 50.9 & 0.912 & 86.7 & 102.2 & \bf\textcolor{blue2}{0.849} & 156.3 & \bf\textcolor{blue2}{0.299}
			\\
			Conv-TT-LSTM~\cite{su2020convolutional} & 53.0 & \bf\textcolor{blue2}{0.915} & \bf \textcolor{red2}{40.5} & 105.7 & 0.840 & \bf\textcolor{blue2}{90.3} & 0.378
			\\
			\hdashline
			 \rule{0pt}{12pt}\bf Proposed Method & \bf\textcolor{red2}{41.5} & \textbf{\textcolor{red2}{0.924}} & \bf\textcolor{blue2}{46.9} &\bf \textcolor{red2}{73.2} & \textbf{\textcolor{red2}{0.879}} & \textbf{\textcolor{red2}{71.6}} & \bf\textcolor{red2}{0.099}\\ 
			\bottomrule
		\end{tabular}
	}
\vspace{-0.1cm}
	\caption{Results on the Moving-MNIST. Higher SSIM values are better while lower MSE and LPIPS values are better. \textbf{\textcolor{red2}{Red}} and \textbf{\textcolor{blue2}{Blue}} indicate the best and the second best, respectively. Ours outperforms the others especially in long-term condition.}
	\label{table1}
	\vspace{0.2cm}
\end{table*}

\begin{figure*}[!t]
	\vspace{-0.0cm}
	\begin{minipage}[b]{1.0\linewidth}
		\centering
		\centerline{\includegraphics[width=17.0cm]{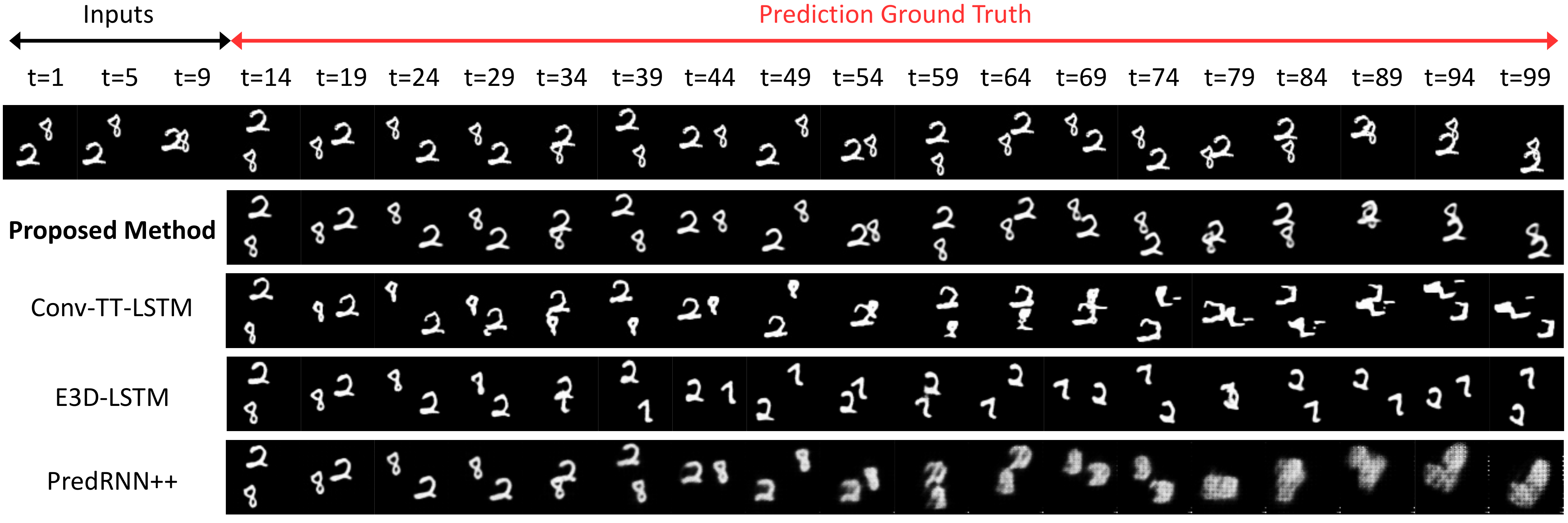}}
	\end{minipage}
		\vspace{-0.6cm}
	\caption{Qualitative results with given 10 frames on the Moving-MNIST. The other results are obtained from official sources.}
	\label{figure3}
		\vspace{-0.05cm}
\end{figure*}

\begin{table*}
	\renewcommand{\arraystretch}{1.0}
	\renewcommand{\tabcolsep}{4.1mm}
	\centering
	\resizebox{0.98\linewidth}{!}{
		\begin{tabular}{l  c c c  c c c  c c }
			\toprule
			\multirow{2}{*}{\bf Prediction Method} & \multicolumn{3}{c}{\bf \makecell{Performance\\(10 $\to$ 20)}} & \multicolumn{3}{c}{\bf \makecell{Performance\\(10 $\to$ 40)}}  & \bf \makecell{Computational Cost\\(10 $\to$ 40)}\\ 
			
			\cmidrule(lr){2-4}
			\cmidrule(lr){5-7}
			   
			& \bf PSNR & \bf SSIM & \bf LPIPS & \bf  PSNR &\bf  SSIM & \bf LPIPS& \bf Inference Time (s) \\
			\bottomrule
			\rule{0pt}{10pt}MCNET~\cite{villegas2017decomposing}~ & 25.95 & 0.804 & - & - & - & -& - \\
			FRNN~\cite{oliu2018folded} & 26.12 & 0.771 & - &23.77  & 0.678& - & -  \\
			PredRNN~\cite{wang2017predrnn} & 27.55 & 0.839 & - & 24.16 & 0.703 & -&- \\
			PredRNN++~\cite{wang2018predrnn++} & \bf\textcolor{red2}{28.62} & 0.888 & 228.9 & \bf\textcolor{blue2}{26.94} & 0.865 & 279.0 & \bf\textcolor{blue2}{0.411}\\
			E3D-LSTM~\cite{wang2018eidetic} &  27.92 & 0.893 & 298.4 & 26.55 & 0.878 & 328.8 & 0.422\\
			Conv-TT-LSTM~\cite{su2020convolutional} & 28.36 & \bf\textcolor{red2}{0.907} & \bf\textcolor{blue2}{133.4} & 26.11 & \bf\textcolor{red2}{0.882} & \bf\textcolor{blue2}{191.2} & 1.188\\
			\hdashline
			\rule{0pt}{12pt}\bf{Proposed Method} &  \bf\textcolor{blue2}{28.61} & \bf\textcolor{blue2}{0.894} & \bf\textcolor{red2}{133.3} &\bf\textcolor{red2}{27.50} & \bf\textcolor{blue2}{0.879} & \bf\textcolor{red2}{159.8} & \bf\textcolor{red2}{{0.147}}\\ \bottomrule
	\end{tabular}}
\vspace{-0.1cm}
	\caption{Results on the KTH. Higher values are better for PSNR and SSIM while lower values are better for LPIPS. \textbf{\textcolor{red2}{Red}} and \textbf{\textcolor{blue2}{Blue}} indicate the best and the second best, respectively. Ours outperforms the others especially in long-term condition.}

	\label{table2}
\end{table*}

\begin{figure*}[t]
	\begin{minipage}[b]{1.0\linewidth}
		\centering
		\centerline{\includegraphics[width=17cm]{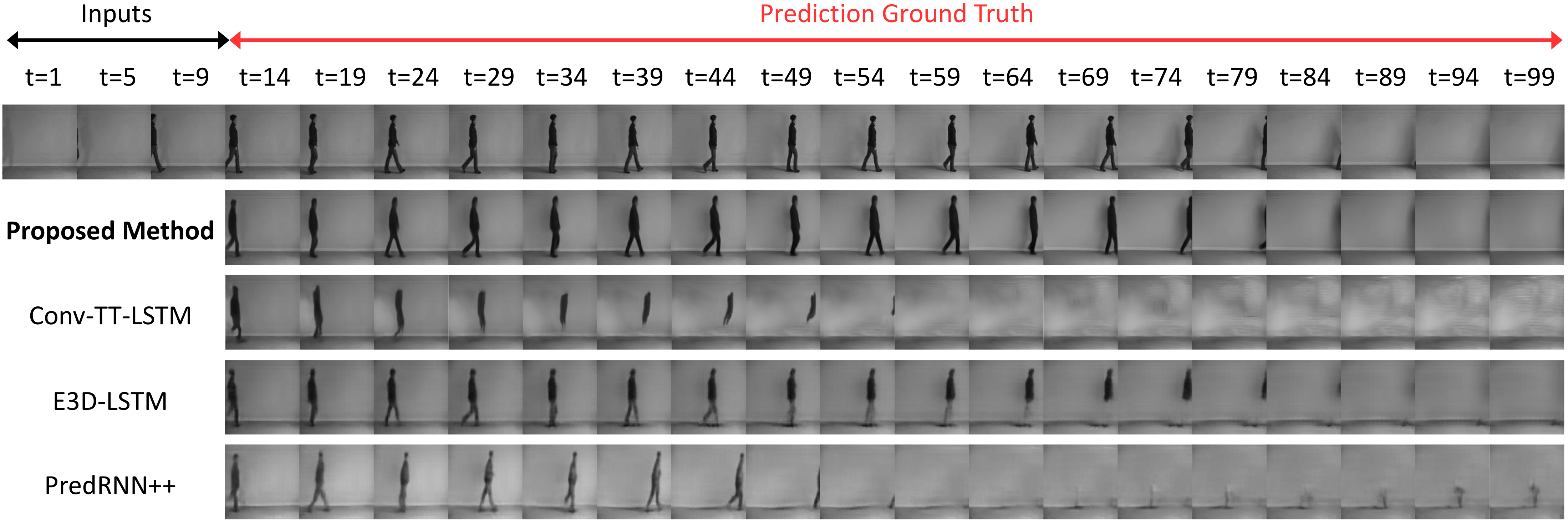}}
	\end{minipage}
		\vspace{-0.6cm}
	\caption{Qualitative results with given 10 frames on the KTH Action. The other results are obtained from official sources.}
	\label{figure4}
\vspace{-0.1cm}
\end{figure*}

\section{Experiments}
\subsection{Datasets}
To validate the proposed method, we utilize both synthetic and natural video datasets. We use a synthetic Moving-MNIST dataset \cite{srivastava2015unsupervised} that is mainly used in video prediction. In addition, we use a KTH Action \cite{schuldt2004recognizing} and a Human 3.6M \cite{ionescu2013human3} datasets including natural videos with human action scenarios.

\noindent{\bf Moving-MNIST.}
The Moving-MNIST \cite{srivastava2015unsupervised} contains the moving of two randomly sampled digits from the original MNIST dataset. Each digit moves in a random direction within a 64$\times$64 size image with a gray scale. The constructed Moving-MNIST dataset consists of 10,000 sequences for training and 5,000 sequences for testing as \cite{wang2018predrnn++}.

\noindent{\bf KTH Action.}
KTH Action dataset \cite{schuldt2004recognizing} consists of 6 types of action videos for 25 subjects. It includes indoor, outdoor, scale variations, and different clothes. Each frame is resized to 128$\times$128 with a gray scale. The videos of 1-16 subjects are used as the training set while the videos of 17-25 subjects are used as the test set.  We follow the experimental setting \cite{villegas2017decomposing} of video prediction for the KTH Action dataset.

\noindent{\bf Human 3.6M.}
The Human 3.6M \cite{ionescu2013human3} includes 17 human action scenarios with total 11 actors. It contains 4 different camera views. Each frame is resized to 64$\times$64 with RGB color channels. Videos of subjects 1, 5, 6, 7, and 8 are used to train the model while videos of subjects 9 and 11 are used to test the model. We follow the experimental setting \cite{villegas2017learning}. 

\subsection{Implementation}
The video frames are normalized to intensity of [0, 1] and resized to 64 $\times$ 64 (MNIST and Human 3.6M) or 128 $\times$ 128 (KTH) as \cite{wang2018predrnn++,villegas2017decomposing,villegas2017learning}. The proposed model is trained by Adam optimizer \cite{kingma2014adam} with a learning rate of 0.0002. Memory slot size $s$ is fixed as 100 for all experiments. Input short-term sequence length $n$ is set as 10. Long-term sequence length $N$ is set as 30 (MNIST) or 40 (KTH and Human 3.6M). Our model is trained to predict corresponding $N$ future frames. We use 4-layer ConvLSTMs for frame prediction.  The overall detailed network structures are described in the supplementary material.

\subsection{Evaluation}
We use MSE, PSNR, SSIM \cite{wang2004image}, and LPIPS \cite{zhang2018unreasonable} to measure the performances. MSE and PSNR are calculated by the pixel-wise difference between the actual frame and the predicted frame. We also evaluate the performance using SSIM that considers the structural similarity between frames. Furthermore, we utilize LPIPS as a perceptual metric, which tends to be similar to the human recognition system \cite{zhang2018unreasonable}. Higher values are better for PSNR and SSIM while lower values are better for MSE and LPIPS. LPIPS results are represented in $10^{-3}$ scale. Single TITAN XP is used to evaluate computational costs for all models. Note that official source codes are used for other methods. 

\noindent{\bf Results on Moving-MNIST.}
Table \ref{table1} shows the performance comparisons with the state-of-the-art methods on the Moving-MNIST. The left part of the table shows the experimental results of 10 frames prediction with the input 10 frames. The right part of the table shows the experimental results for 30 frames prediction with 10 input frames. The proposed method outperforms the other state-of-the-art methods. In particular, our method far surpasses the others in predicting 30 frames in terms of the LPIPS metric. In addition, the proposed method shows much better results on the computational cost compared to the other methods. Compared to other complex RNN-based methods, we adopt simple ConvLSTMs. Further, memory feature $F^{mem}$ is extracted only once at the beginning, which is advantageous in computational cost. Figure \ref{figure3} shows examples of frames predicted by the proposed method and other video prediction methods. As shown in the figure, the predicted frames of the proposed method show convincingly similar results to the ground truth. However, the prediction results by other methods show that they lose the trajectories or the shape of digits, especially in long-term condition.

\begin{table}[t!]{
		\renewcommand{\arraystretch}{1.25}
		\renewcommand{\tabcolsep}{0.9mm}
		\centering
		\resizebox{0.999\linewidth}{!}{
			\begin{tabular}{l c c c c c c }
				\toprule
				\multirow{2}{*}{\bf Prediction Method \;} &\multicolumn{3}{c}{\bf \makecell{Performance \\(10 $\to$ 40)}}  &\multicolumn{3}{c}{\bf \makecell{Performance \\(10 $\to$ 40, Last 10)}}\\ 
				\cmidrule(lr){2-4}
				\cmidrule(lr){5-7}
				
				& \bf PSNR & \bf SSIM& \bf LPIPS & \bf PSNR & \bf SSIM& \bf LPIPS\\ \bottomrule
				\rule{0pt}{10pt}\makecell[l]{PredRNN++ \cite{wang2018predrnn++}} & 23.23   & 0.876 &  106.6 & 22.10   & 0.862 &  123.3 \\ \rule{0pt}{0pt}\makecell[l]{E3D-LSTM \cite{wang2018eidetic}} & 22.33 & 0.850 &  113.3 & 21.11 & 0.825 &  140.1 \\ \hdashline
				\rule{0pt}{12pt}\bf Propose Method & \bf 24.97 &\bf 0.919 &  \bf 63.2 & \bf 23.46 &\bf 0.902 &  \bf 80.1\\
				\bottomrule
		\end{tabular}}
		\vspace{-0.1cm}
		\captionof{table}{Results on the Human 3.6M. Higher PSNR and SSIM are better while lower LPIPS is better.}
		\label{table3}}
	\vspace{-0.25cm}
	
\end{table}

\begin{figure}[t]
	\begin{minipage}[b]{1.0\linewidth}
		\centering
		\centerline{\includegraphics[width=8.5cm]{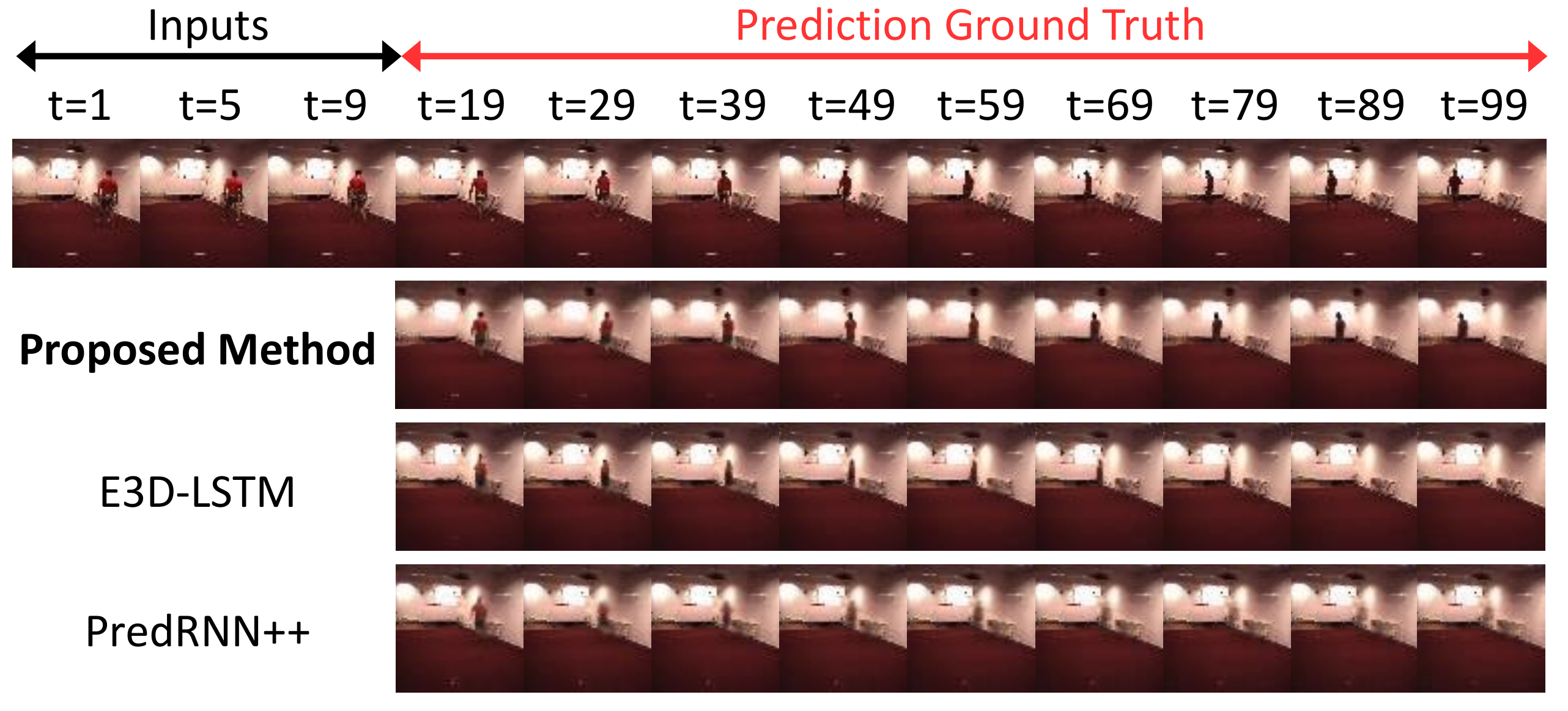}}
	\end{minipage}
		\vspace{-0.6cm}
	\caption{Qualitative results with given 10 frames on the Human 3.6M. Results of the others are from official sources.}
	\label{figure5}
		\vspace{-0.25cm}
\end{figure}

\noindent{\bf Results on KTH Action.}
Table \ref{table2} shows the quantitative results of the proposed method and other state-of-the-art methods on the KTH Action dataset. The left part of the table shows the experimental results for predicting 20 frames with 10 input frames. The right part of the table indicates performances for predicting the next 40 frames. As shown in the table, the proposed method mostly surpasses the other state-of-the-art methods in predicting 40 frames. Especially, it is significant in the human perceptual metric (\textit{i.e.}, LPIPS). In addition, the proposed method shows a much faster inference speed compared to the other methods also on the KTH. Figure \ref{figure4} shows qualitative long-term prediction results for input sequence with limited dynamics on the KTH. This input motion is limited because motion actually starts from the middle. As shown in the figure, the other methods fail to capture the detailed leg movement, especially in long-term condition. Whereas, our predicted frames are very similar to the ground truth frames. The proposed method maintains a clear shape of the legs while following the long-term trajectories even in such a challenging condition (\textit{i.e.}, limited dynamics). 

\noindent{\bf Results on Human 3.6M.}
Table \ref{table3} shows the performance comparisons with the other methods on the Human 3.6M. The left part of the table shows the experimental results for predicting 40 frames with given 10 input frames. The right part of the table indicates performances for the last 10 frames among the future 40 frames. The proposed method outperforms other state-of-the-art video prediction methods both in predicting 40 future frames and predicting the last 10 frames. Figure \ref{figure5} shows qualitative results for the long-term prediction on the Human 3.6M. The proposed method captures direction changing in the long-term while the other methods show the disappearance of a person at the corner. Compared to the others, the proposed method properly captures the long-term motion context with redirection.

\begin{table}[t]{
		\renewcommand{\arraystretch}{1.3}
		\renewcommand{\tabcolsep}{0.0mm}
		\centering
		\resizebox{0.999\linewidth}{!}{
			\begin{tabular}{l  c  c  c  c }
				\toprule
				\multirow{2}{*}{\bf Prediction Method} &\multicolumn{3}{c}{\bf \makecell{Performance \\(10 $\to$ 40, Last 10)}}  & \bf \makecell{Computational Cost\\(10 $\to$ 40)}\\ 
				\cmidrule(lr){2-4}
				& \bf PSNR & \bf \ SSIM \ & \bf LPIPS & \bf Inference Time (s)\\ \bottomrule
				\rule{0pt}{15pt}Model w/o LMC-Memory & 25.29   & 0.851 &  321.1 & 0.118\\
				\rule{0pt}{20pt}\makecell[l]{Model w/ LMC-Memory \\ (Non-local Motion Context)\; } & 25.57 & 0.854 &  298.8 & 0.136 \\ \hdashline
				\rule{0pt}{18pt}\makecell[l]{\bf Model w/ LMC-Memory \\ \bf (Local Motion Context) } & \bf 26.21 & \bf 0.862 & \bf 195.6 & 0.147\\
				\bottomrule
		\end{tabular}}
			\vspace{-0.1cm}
		\captionof{table}{Effects of the network designs on the performance and the computational cost. Performance evaluations are conducted on KTH Action dataset.}
		\label{table4}}
	\vspace{-0.1cm}
\end{table}

\begin{figure}[t]
	\begin{minipage}[b]{1.0\linewidth}
		\centering
		\centerline{\includegraphics[width=8.5cm]{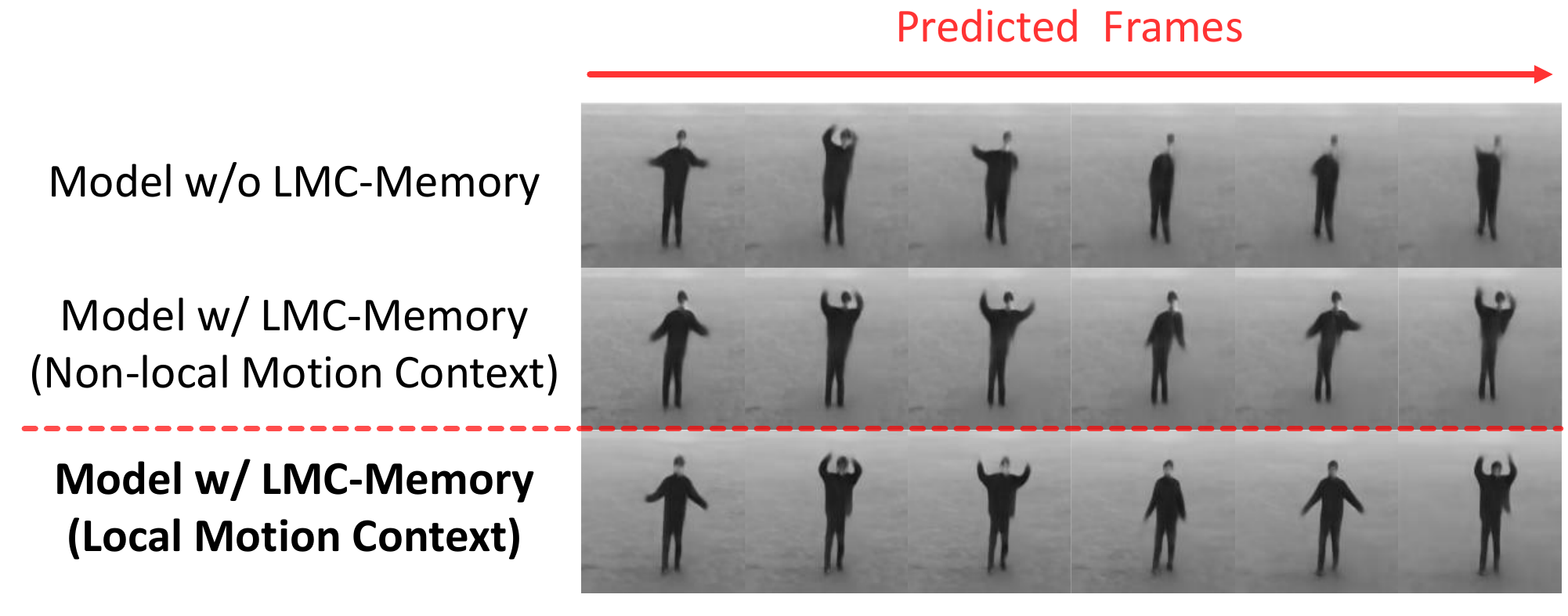}}
	\end{minipage}
		\vspace{-0.6cm}
	\caption{Video prediction qualitative results for the different network designs on the KTH Action dataset.}
	\label{figure6}
	\vspace{-0.3cm}
\end{figure}

\subsection{Ablation Study}
We analyze the effects of network designs by ablating them as shown in Table \ref{table4}. In detail, we investigate the effectiveness of the LMC-Memory (\textit{i.e.}, memory alignment learning) and the local motion context (\textit{i.e.}, memory query decomposition). The baseline, \textit{`Model w/o LMC-Memory}' consists of the spatial encoder, the ConvLSTMs, and the frame decoder. The second one, `\textit{Model w/ LMC-Memory (Non-local motion context)}' contains LMC-Memory but it does not adopt memory query decomposition to use local motion context as memory queries. This model uses a globally pooled motion context feature as a query. The last one indicates our final proposed model of this paper. As shown in the table, each component contributes to the performances in predicting the last 10 among 40 frames. The final model outperforms the other models, especially in terms of perceptual metric LPIPS. These results show that locally manipulated query boosts the effects of the memory since it is more accessible to store and recall the motion context with low-dimensional dynamics. Further, the additional computational cost to use the LMC-Memory is marginal. 

Figure \ref{figure6} shows qualitative results for different network designs. The first model does not properly capture the long-term motion. The second one predicts long-term motion to some extent. However, the detailed local parts are distorted because it addresses the motion context in an only global manner. The final model effectively predicts future frames by properly capturing the context of long-term motion.

\begin{figure}[t]
	\begin{minipage}[b]{1.0\linewidth}
		\centering
		\centerline{\includegraphics[width=8.5cm]{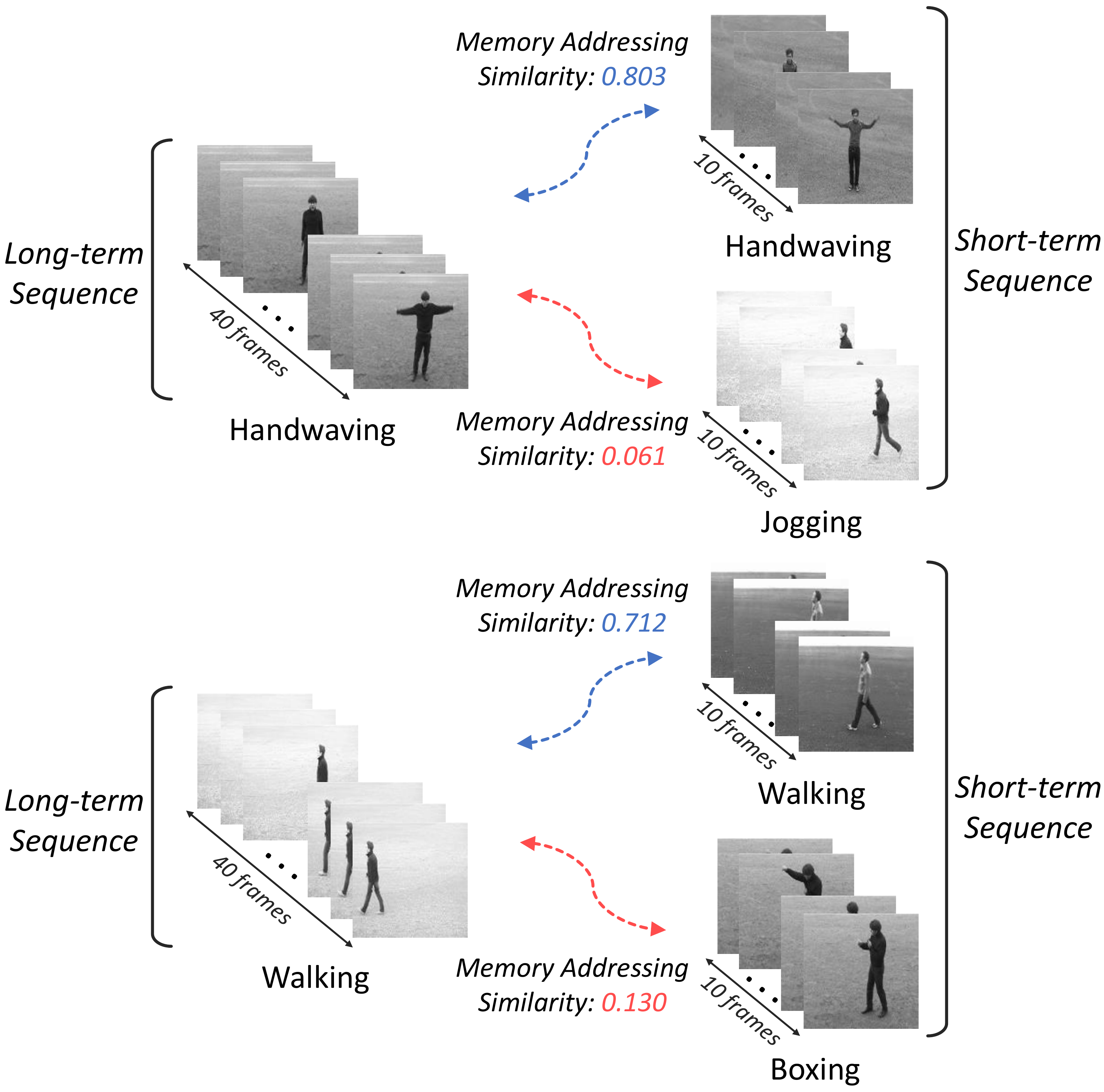}}
	\end{minipage}
		\vspace{-0.6cm}
	\caption{Examples of similarity between memory addressing vectors from long-term and short-term sequences in the KTH Action dataset.}
	\label{figure7}
		\vspace{-0.30cm}
\end{figure}

\subsection{Memory Addressing}
We analyze the memory addressing for different sequences. Figure \ref{figure7} shows the cosine similarity values between addressing vectors from long-term and short-term sequences including different subjects. The areas addressed in the memory are more comparable (similarity between addressing vectors is high) in the case of the same action scenario than in the case of the different actions. It shows that the long-term and short-term features that belongs to the similar action are convincingly aligned in the memory.

\section{Conclusion}
The objective of the proposed work is to predict future frames being aware of the long-term motion context. To this end, we propose the LMC-Memory with the alignment learning scheme to effectively store abundant long-term contexts of training data and recall suitable motion context even from limited inputs. In addition, we utilize memory query decomposition to separate overall motion into low-dimensional dynamics. It enables to cope with the high-dimensionality in terms of utilizing motion contexts in the memory. As a result, the proposed method outperforms the state-of-the-art methods with sophisticated RNNs. In particular, it is significantly noticeable in long-term condition. Further, the effectiveness of the proposed method is analyzed in both quantitative and qualitative ways. \vspace{0.17cm}\\
\noindent{\bf Acknowledgement.}
This work was partly supported by the IITP grant (No.\ 2020-0-00004) and BK 21 Plus project.

{\small
\bibliographystyle{ieee_fullname}
\bibliography{refs}
}

\end{document}